\documentclass[manuscript,screen]{acmart}
\usepackage{subfigure}
\usepackage{multirow}

\usepackage{graphicx}
\usepackage{bm}
\usepackage{amsfonts}

\usepackage{amssymb}
\usepackage{amsmath}
\usepackage{natbib}
\usepackage{threeparttable}

\usepackage{mathrsfs}
\usepackage{url}
\usepackage{caption}
\usepackage{times}

\AtBeginDocument{%
  }
\setcopyright{acmcopyright}
\copyrightyear{2022}
\acmYear{2022}
\acmDOI{XXXXXXX.XXXXXXX}
\renewcommand\footnotetextcopyrightpermission[1]{}

\acmJournal{TKDD}
\acmVolume{0}
\acmNumber{0}
\acmArticle{111}
\acmMonth{6}
\begin{document}

\title{TPRNN: A Top-Down Pyramidal Recurrent Neural Network for Time Series Forecasting}

\author{Ling Chen}
\authornote{The corresponding author.}
\affiliation{%
\department{College of Computer Science and Technology}
  \institution{Zhejiang University}
  \streetaddress{38 Zheda Road}
  \city{Hangzhou}
  \state{Zhejiang}
  \country{China}
  \postcode{310027}
  }
\email{lingchen@cs.zju.edu.cn}

\author{Jiahua Cui}
\affiliation{%
  \department{College of Computer Science and Technology}
  \institution{Zhejiang University}
  \streetaddress{38 Zheda Rd}
  \city{Hangzhou}
  \state{Zhejiang}
  \country{China}
  \postcode{310027}}
\email{15940349242@163.com}

\begin{abstract}
  Time series refer to a series of data points indexed in time order, which can be found in various fields, e.g., transportation, healthcare, and finance. Accurate time series forecasting can enhance optimization planning and decision-making support. Time series have multi-scale characteristics, i.e., different temporal patterns at different scales, which presents a challenge for time series forecasting. In this paper, we propose TPRNN, a \textbf{T}op-down \textbf{P}yramidal \textbf{R}ecurrent \textbf{N}eural \textbf{N}etwork for time series forecasting. We first construct subsequences of different scales from the input, forming a pyramid structure. Then by executing a multi-scale information interaction module from top to bottom, we model both the temporal dependencies of each scale and the influences of subsequences of different scales, resulting in a complete modeling of multi-scale temporal patterns in time series. Experiments on seven real-world datasets demonstrate that TPRNN has achieved the state-of-the-art performance with an average improvement of 8.13\% in MSE compared to the best baseline.
\end{abstract}





\maketitle

\section{Introduction}
With the development of society and advances in technology, there is an increasing amount of time series data generated in our daily life\cite{TKDD1,TKDD2,TKDE1}. Time series forecasting, which predicts future data values by analyzing previously observed time series, has rapidly been used in many important real-world applications, e.g., transportation planning \cite{MRA-BGCN}, energy consumption \cite{DB-Net}, and weather prediction \cite{Airformer}.

Time series exhibit different patterns at different scales, presenting a significant challenge for time series forecasting models. For example, in traffic, time series recorded with a sampling rate of one-hour mainly reflect local patterns (e.g., daily patterns). After changing to a sampling rate of one-day, it is easier to capture global patterns (e.g., monthly patterns or yearly patterns). A good time series forecasting model should be capable of modeling multi-scale temporal patterns in time series.

Early time series forecasting models can be divided into two categories: Statistical models and traditional machine learning models. Statistical models, e.g., Auto-Regressive Integrated Moving Average (ARIMA) \cite{ARMA} and Exponential Smoothing (ES) \cite{ES}, mainly use statistical methods to analyze time series, often assuming that time series are stable and have linear dependencies. However, time series often contain complex, non-linear dependencies that statistical models cannot handle. Traditional machine learning models, e.g., Support Vector Regression (SVR) \cite{SVR}, Bayesian Model \cite{box2015time}, and Hidden Markov Model (HMM) \cite{HMM}, can model non-linear temporal dependencies in time series, but they require complex feature engineering before forecasting. Time series in different domains require different feature engineering designs, which limit the applicability of these models.

Recently, with the development of deep learning, many studies have attempted to use deep learning models to model complex dependencies in time series, e.g., Recurrent Neural Networks (RNNs) \cite{MiRNN}, Temporal Convolutional Networks (TCNs) \cite{TCN,TKDD3}, and Transformers \cite{LogTrans,Informer,Autoformer,FEDformer,ETC,Pyraformer,Triformer}. Compared with statistical models and traditional machine learning models, deep learning models can automatically learn the underlying features of data with carefully designed network structures, eliminating the need for complex feature engineering. To model the multi-scale temporal patterns in time series, some studies \cite{TCN,TAMS-RNN,NHITS} attempt to construct subsequences of different scales from the original time series and model them independently, but they neglect the influences of subsequences of different scales. Considering the values in subsequences as nodes, some recent studies \cite{ETC,Pyraformer,Triformer} stack these subsequences into a pyramid structure and define intra-scale and inter-scale edges, which can model the temporal dependencies within each scale and the influences of subsequences of different scales. However, these models only capture the interactions between nodes and their neighbors within a small range, neglecting the influences from distant nodes and the entire subsequence, leading to insufficient modeling of the influences of subsequences of different scales.

To address the above problems, we propose TPRNN, a \textbf{T}op-down \textbf{P}yramidal \textbf{R}ecurrent \textbf{N}eural \textbf{N}etwork for time series forecasting. From the largest scale at the top of the pyramid to the smallest scale at the bottom, TPRNN aggregates the global information of each scale before passing it to all nodes in the adjacent lower level of the pyramid. This enables subsequences to incorporate the influence of adjacent larger scale subsequences while modeling their own scale's temporal dependencies, thereby achieving comprehensive modeling of multi-scale temporal patterns in time series.
Our main contributions are summarized as follows:
    \begin{itemize}
    \item Propose an inter-scale interaction block to extract global information from subsequences and integrate it into the modeling of temporal dependencies of the adjacent smaller scale subsequence, which can capture the influences of subsequences of different scales and avoid the high computational complexity of directly modeling the interactions of nodes of different scales.
    \item Propose an intra-scale interaction block to model the subsequences of each scale and selectively filter information through gating mechanisms, which can capture the temporal dependencies within each scale.
    \item Conduct extensive experiments on seven real-world datasets. The experiment results demonstrate that TPRNN achieves competitive performance, with an average improvement of 8.13\% in MSE compared to the best baseline.
    \end{itemize}

\section{Related Work}

\subsection{Time Series Forecasting}
Time series forecasting models can be divided into three categories: Statistical models, traditional machine learning models, and deep learning models. Statistical models (e.g., ARIMA and ES) predict the future by statistically analyzing historical data, but require time series to be stationary or stationary after differencing. Machine learning models (e.g., SVR, Bayesian Model, and HMM) have stronger modeling capabilities, but their applicability is limited due to the requirement for complex and customized feature engineering.

Deep learning models have achieved significant success in time series forecasting, leveraging their powerful modeling capabilities. RNNs have strong sequence modeling capabilities, making them highly popular in time series forecasting. For example, DeepAR \cite{DeepAR} uses an autoregressive recurrent network to produce probabilistic forecasts. ES-RNN \cite{ES-RNN} combines the power of exponential smoothing methods and RNNs to effectively capture the main components of time series, including seasonality and non-linear trends. LSTNet \cite{LSTNet} uses recurrent-skip connections and convolutions to capture both long-term and short-term dependencies in time series. However, the recurrent-based models are not parallelizable, and they also suffer from the vanishing gradient problem, which makes them ineffective in handling long-term dependencies. Some works \cite{TCN,weather} introduce dilated casual convolutions to extract features from time series. Limited to the size of the convolutional kernel, these models can only focus on local features and are also hard to capture long-term dependencies in time series. In addition, a recent MLP-based model DLinear \cite{DLinear} decomposes the input into seasonal and trend components and employs fully connected layers to model each component separately, which achieves excellent performance.

Recently, transformer-based models demonstrate powerful capabilities in many fields, e.g., natural language processing \cite{ETC} and computer vision \cite{fan2021multiscale}. The powerful modeling capabilities of self-attention mechanism for sequential data result in the development of numerous transformer-based models for time series forecasting. LogTrans \cite{LogTrans} uses causal convolution to enhance the local information and log-sparse attention to reduce the computational complexity. Informer \cite{Informer} uses a KL-divergence-based method to select dominant queries, which also significantly reduces the computational complexity. Autoformer \cite{Autoformer} applies seasonal-trend decomposition to split time series into two parts and uses auto-correlation mechanism instead of self-attention to focus on the connections of subsequences in time series. FEDformer \cite{FEDformer} proposes the mixture of experts’ strategies to decompose time series and calculates attention value between different frequency components instead of different time points to model the global patterns in time series.

However, the aforementioned models are limited to analyzing and modeling data only at the original scale, without fully utilizing the multi-scale characteristics of time series. Modeling time series at a single scale may result in the mixture of patterns.

\subsection{Multi-Scale Neural Networks}
With the development of deep learning, many multi-scale neural networks have been proposed. The key distinction between multi-scale neural networks and other neural networks lies in their ability to model data at different scales and capture both intra-scale and inter-scale interactions. Multi-scale neural networks have been widely applied in various fields, e.g., computer vision, natural language processing, and time series forecasting. Multi-scale ViT \cite{fan2021multiscale} captures both global and local dependencies in image data by connecting the seminal idea of multi-scale feature hierarchies with transformer models. ETC \cite{ETC} introduces several global nodes to model the intra-scale and inter-scale information interactions in natural language.

In time series forecasting, time series show different patterns at different scales, and recently many multi-scale neural networks have been proposed to model the multi-scale temporal patterns in time series. For example, MiRNN \cite{MiRNN} utilizes a deep wavelet decomposition network to obtain subsequences of different scales and uses three strategies (truncation, initialization, and message passing) to capture the influences of different subsequences. Pyraformer \cite{Pyraformer} extends the global-local structure of ETC \cite{ETC} into a pyramid structure, and calculates attention values between nodes and their neighbors (including neighboring nodes, parent node, and child nodes). However, these models either fail to model the inter-scale interactions of different scales or only capture the interactions among adjacent nodes, neglecting the interactions with other nodes or the entire subsequence.

\section{Methodology}
\subsection{Problem Definition}
We first provide the definition of time series forecasting. Given the input sequence $\bm{X}_{1:T}=\{\bm{x}_t|\bm{x}_t\in \mathbb{R}^D,t\in[1:T]\}$, where $\bm{x}_t$ is the values at the time step $t$, $T$ is the input length, and $D$ is the feature dimension. The time series forecasting task is as follows:
\begin{equation}
{\hat{\bm{Y}}}_{T+1:T+H}=f(\bm{X}_{1:T},\theta),
\end{equation}
where ${\hat{\bm{Y}}}_{T+1:T+H}$ is the prediction, $H$ is the prediction length, $f$ is the forecasting model, and $\theta$ is the parameters of the forecasting model.

\subsection{Overview}
\begin{figure*}[t]
\centering
\includegraphics[width=0.85\textwidth]{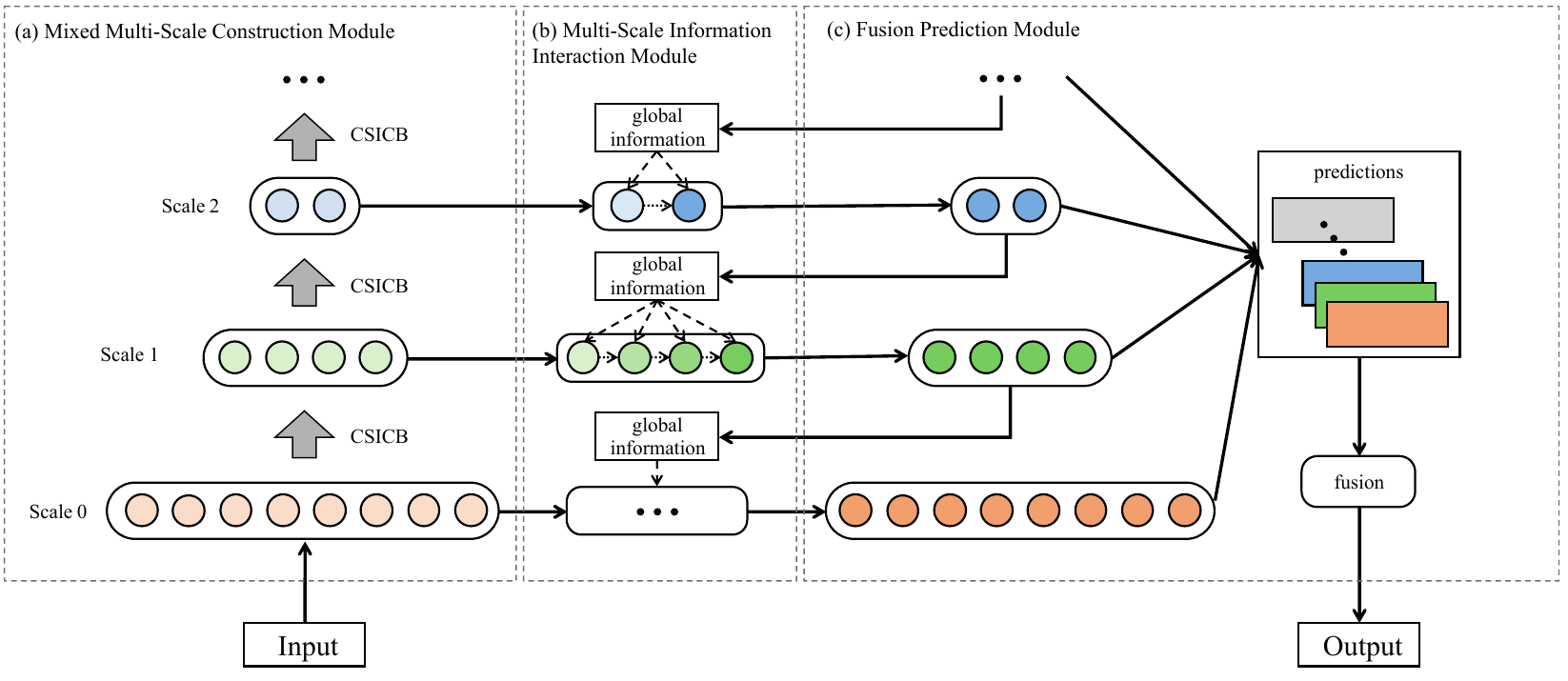}
\caption{The framework of TPRNN. (Best viewed in color).
}
\label{fig:framework}
\end{figure*}

Fig. \ref{fig:framework} shows the framework of TPRNN, which consists of three parts: (a) The mixed multi-scale construction module transforms input time series from smaller scale to larger scale hierarchically and forms a pyramid structure; (b) The multi-scale information interaction module, composed of an intra-scale interaction block and an inter-scale interaction block, is utilized to model both the temporal dependencies of time series in each scale and the influences of subsequences of different scales in a top-down manner within the pyramid structure; (c) The fusion prediction module is utilized to generate predictions at different scales and then combines these predictions to obtain the final prediction. The details of these modules are described in the following subsections.

\subsection{Mixed Multi-Scale Construction Module}
\begin{figure}[t]
\centering
\includegraphics[width=0.50\columnwidth]{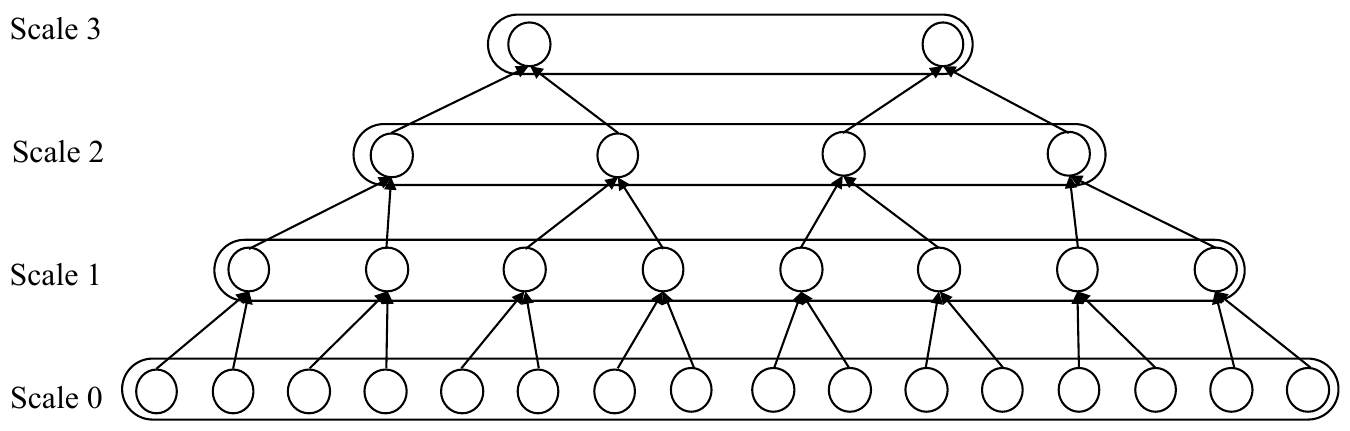}
\caption{A pyramid structure consisting of four subsequences, where the bottom layer represents the input sequence, and the other layers are subsequences obtained through multi-scale construction methods.}
\label{fig:pyramid_structure}
\end{figure}

Many works analyze multi-scale temporal patterns in time series using a pyramid structure. Before introducing the mixed multi-scale construction module in TPRNN, we give a brief introduction to the pyramid structure in time series forecasting. Given a time series $\bm{X}\in\mathbb{R}^{L\times D}$, we can extract multi-scale features through different multi-scale construction methods and obtain subsequences of different scales. The set of these subsequences is $\bm{\mathcal{X}}=\left\{\bm{X}^0,\cdots,\bm{X}^s,\cdots,\bm{X}^C\right\}$, where $C$ is the number of subsequences of different scales, $\bm{X}^0$ is the original input, $\bm{X}^s\in\mathbb{R}^{L_s\times D}$ is the $s$-th subsequence, $L_s$ is the length of the $s$-th subsequence. Since multi-scale construction methods reduce the length of the input sequence while extracting features, stacking subsequences of different scales together based on their scales results in a pyramid structure. We refer to values in the input sequence or feature values in subsequences of different scales as nodes in the pyramid structure. A pyramid structure consisting of four subsequences is illustrated in Fig. \ref{fig:pyramid_structure}.

\begin{figure*}[t]
\centering
\includegraphics[width=0.85\columnwidth]{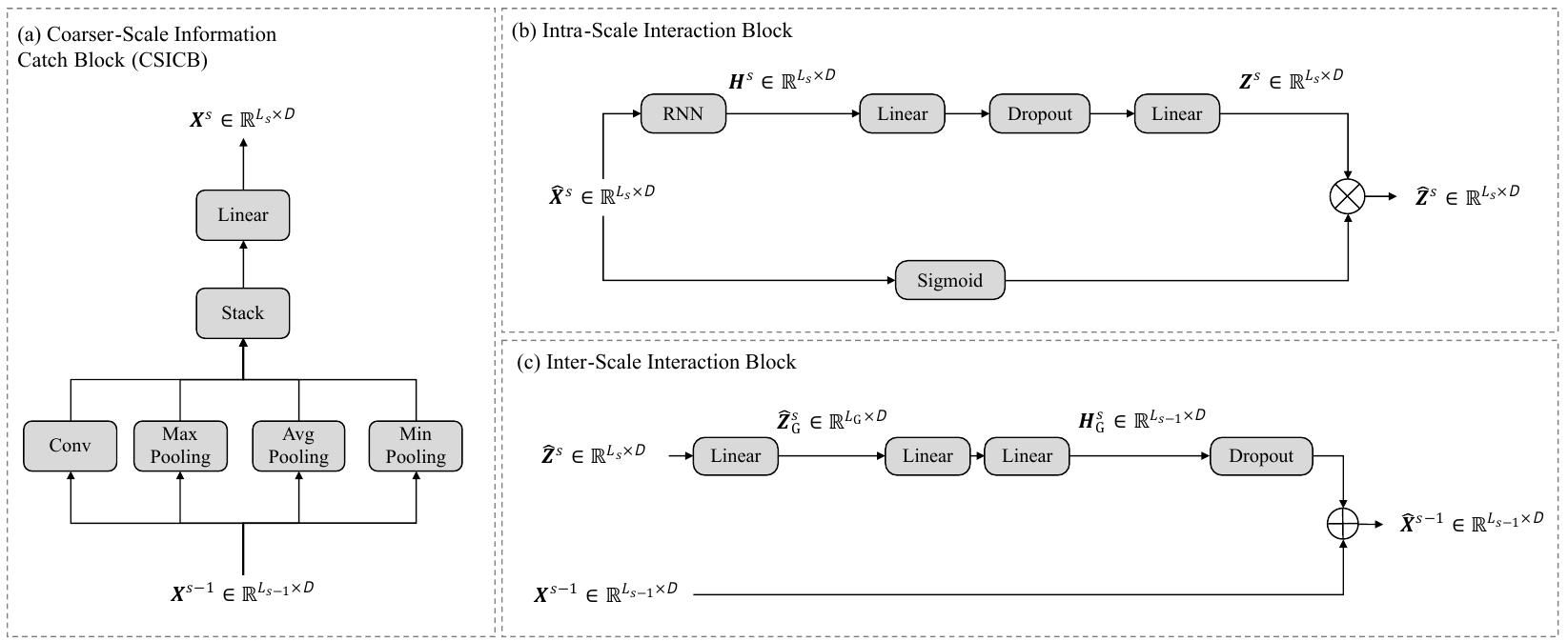}
\caption{An illustration of the main components of TPRNN. (a) The structure of the CSICB, consists of four multi-scale construction methods and one linear layer. (b) The intra-scale interaction block consists of RNN, gating units, two linear layers, and one dropout layer. (c) The inter-scale interaction block consists of three linear layers composing a bottleneck structure and a dropout layer.}
\label{fig:main_components}
\end{figure*}

Existing works often adopt a single multi-scale construction method when building a pyramid structure. This means that subsequences are solely composed of a specific type of information from the input, neglecting other types of information and hindering the capture of temporal patterns in time series. Therefore, we propose a mixed multi-scale construction module that combines multiple multi-scale construction methods to capture different types of information from the input. Within this module, we use the coarser-scale information catch block (CSICB) to extract multi-scale features. A CSICB consists of four multi-scale construction methods: convolution, average pooling, max pooling, and min pooling. The convolution method utilizes the backpropagation algorithm to automatically update the parameters of the convolutional kernels, allowing it to adaptively capture the most valuable features from the input. In addition, average pooling, max pooling, and min pooling methods extract statistical information that is difficult to obtain through convolution. Finally, different weights are assigned to the four types of features through a linear layer. The structure of the CSICB is illustrated in Fig. \ref{fig:main_components}(a) and the implementation is as follows:
\begin{flalign}
    \begin{aligned}
        &\ \bm{X}_{\rm H}^s = [\mathrm{Conv}\left(\bm{X}^{s-1}\right),\mathrm{MaxPooling}\left(\bm{X}^{s-1}\right), &\\
        &\ \mathrm{MinPooling}\left(\bm{X}^{s-1}\right),\mathrm{AvgPooling}\left(\bm{X}^{s-1}\right)] &\\
        &\ \bm{X}^s = \mathrm{Linear}\left(\mathrm{Stack}\left(\bm{X}_{\rm H}^s\right)\right), &
    \end{aligned}
\end{flalign}
where $s\in\left[1,C\right]$, and $\bm{X}_{\rm H}^s$ is the intermediate state. $\rm Stack\left(\cdot\right)$ and $\rm Linear\left(\cdot\right)$ are the stacking operation and linear transformation, respectively.
\subsection{Multi-Scale Information Interaction Module}

After obtaining a set of subsequences of different scales, it is necessary to further facilitate information interaction in order to achieve a comprehensive modeling of multi-scale temporal patterns. Specifically, multi-scale temporal patterns can be divided into two parts: The temporal dependencies within a single-scale subsequence and the influences of subsequences of different scales. In the multi-scale information interaction module, we design two blocks: intra-scale interaction block and inter-scale interaction block. They respectively model two parts of multi-scale temporal patterns. Within the pyramid structure, TPRNN utilizes these two blocks in an alternating fashion, sequentially modeling subsequences of different scales from top to bottom.
\paragraph{\textbf{Intra-Scale Interaction Block}}
The intra-scale interaction block is used to model the temporal dependencies within a single-scale subsequence. The structure of this block is illustrated in Fig. \ref{fig:main_components}(b). The RNN is employed to capture the temporal dependencies, while the gating mechanisms are responsible for controlling and filtering information at different positions. In TPRNN, RNN of the intra-scale interaction block can be vanilla RNN, LSTM, GRU, and bi-directional LSTM, etc.

Assuming that the subsequence input to the intra-scale interaction block of the $s$-th scale is ${\hat{\bm{X}}}^s=\{{\hat{\bm{x}}}_1^s,{\hat{\bm{x}}}_2^s,\cdots,{\hat{\bm{x}}}_{L_s}^s\}\in\mathbb{R}^{L_s\times D}$, where $s\in\left[0,C\right]$. The multi-scale information interaction module alternates between the intra-scale interaction block and the inter-scale interaction block. When $s$ is equal to $C$, the input subsequence ${\hat{\bm{X}}}^C$ of the intra-scale interaction block is $\bm{X}^C$, which is obtained from the mixed multi-scale construction module. When $s$ is not equal to $C$, the input of the intra-scale interaction block is ${\hat{\bm{X}}}^s$, which is the result of the inter-scale interaction block at the $\left(s+1\right)$-th scale. The overall process is as follows:
\begin{equation}
\begin{split}
    \bm{H}^s &= {\rm RNN}\left({\hat{\bm{X}}}^s\right) \\
    \bm{Z}^s &= {\rm Linear}_2^{\rm ita}\left({\rm Dropout}\left({\rm Linear}_1^{\rm ita}\left(\bm{H}^s\right)\right)\right), \\
\end{split}
\end{equation}
where $\bm{H}^s$ is the hidden state of RNN and ${\rm Linear}_1^{\rm ita}\left(\cdot\right)$ is the first linear layer in the intra-scale interaction block, which transforms the features of the input sequence ${\hat{\bm{X}}}^s$ into a high-dimensional space and combines the features across different channels. After passing through ${\rm Dropout}(\cdot)$, the data are then transformed back to its original dimension using ${\rm Linear}_2^{\rm ita}(\cdot)$, which is the second linear layer in the intra-scale interaction block.

After passing through RNN and linear layers, we further introduce gating mechanisms to control the subsequence $\bm{Z}^s$. Since gating mechanisms can effectively control the information propagation, we utilize ${\hat{\bm{X}}}^s$ as a “gate” to control the information in $\bm{Z}^s$:
\begin{equation}
    {\hat{\bm{Z}}}^s={\rm Sigmoid}\left({\hat{\bm{X}}}^s\right)\circ\bm{Z}^s,
\end{equation}
where ${\hat{\bm{Z}}}^s\in\mathbb{R}^{L_s\times D}$ is the subsequence of the $s$-th scale after intra-scale interaction, ${\rm Sigmoid}\left(\cdot\right)$ is the sigmoid activation function to generate the update gate, and $\circ$ is the Hadamard product. The reason for introducing gating mechanisms to control the output of RNN is twofold. Firstly, it allows for direct control of the network output using input sequence ${\hat{\bm{X}}}^s$, which helps to balance the potential errors that may arise from considering historical data in RNN. Secondly, it provides a more direct pathway during the gradient backpropagation process, resulting in higher network training efficiency. Through the intra-scale interaction block, the subsequence of the $s$-th scale is updated from ${\hat{\bm{X}}}^s$ to ${\hat{\bm{Z}}}^s$.

\paragraph{\textbf{Inter-Scale Interaction Block.}}
When modeling the influences of subsequences of different scales, directly considering the interactions between nodes at two scales would lead to an excessive computational complexity $O\left(L^2\right)$, where $L$ is the input length. In addition, the influences of subsequences of different scales are typically unidirectional, meaning that a subsequence is influenced by the larger scale subsequences but not by the smaller scale subsequences.

To address the aforementioned issues, we introduce the inter-scale interaction block to capture the influences of subsequences of different scales. The structure of this block is illustrated in Fig. \ref{fig:main_components}(c), which consists of three linear layers and one dropout layer. The first linear layer extracts global information ${\hat{\bm{Z}}}_{\rm G}^s\in\mathbb{R}^{L_{\rm G}\times D}$ from the subsequence of the $s$-th scale, where $L_{\rm G}$ is the length of the global information and is much smaller than $L_s$ or $L_{s-1}$. The second linear layer facilitates the interactions of global information across different dimensions, enhancing the expressive power of the global information. The third linear layer maps the updated global information to $L_{s-1}$, obtaining the influence of the subsequence of the $s$-th scale $\bm{H}_{\rm G}^s$ on the subsequence of the $(s-1)$-th scale $\bm{X}^{s-1}$. After passing through a dropout layer, the updated influence is added to $\bm{X}^{s-1}$, enabling the modeling of the influence between two scales.

By leveraging global information, we can ensure that irrelevant information or noise is filtered out, and only important information is considered. The overall process is as follows:
\begin{equation}
    \begin{split}
        {\hat{\bm{Z}}}_{\rm G}^s &= {\rm Linear}_1^{\rm ite}({\hat{\bm{Z}}}^s) \\
        \bm{H}_{\rm G}^s &= {\rm Linear}_3^{\rm ite}\left({\rm Linear}_2^{\rm ite}\left({\hat{\bm{Z}}}_{\rm G}^s\right)\right) \\
        {\hat{\bm{X}}}^{s-1} &= {\rm Dropout}\left(\bm{H}_{\rm G}^s\right)+\bm{X}^{s-1},	\\
    \end{split}
\end{equation}
where ${\rm Linear}_1^{\rm ite}(\cdot)$, ${\rm Linear}_2^{\rm ite}(\cdot)$, and ${\rm Linear}_3^{\rm ite}(\cdot)$ are three linear layers composing a bottleneck structure in the inter-scale interaction block.

\subsection{Fusion Prediction Module}
To fully utilize subsequences of different scales for prediction, we propose the fusion prediction module as the predictor of TPRNN. The fusion prediction module consists of individual predictors for each scale and a fusion layer. The predictors for each scale are constructed using a linear layer, which takes ${\hat{\bm{Z}}}^s$ as input and generates prediction ${\hat{\bm{Z}}}_{\rm p}^s\in\mathbb{R}^{H\times D}$ at the $s$-th scale:
\begin{equation}
    {\hat{\bm{Z}}}_{\rm p}^s={\rm Linear}\left({\hat{\bm{Z}}}^s\right).
\end{equation}

We employ a linear layer to combine the predictions of different scales and obtain the final prediction $\hat{\bm{Y}}$:
\begin{equation}
    \hat{\bm{Y}}={\rm Linear_F}\left({\rm Stack}\left({\hat{\bm{Z}}}_{\rm p}^0,\cdots,{\hat{\bm{Z}}}_{\rm p}^C\right)\right),
\end{equation}
where ${\rm Linear_F}\left(\cdot\right)$ is a linear layer without bias and ${\rm Stack}\left(\cdot\right)$ is the stacking operation.

\section{Experiments}

\subsection{Datasets and Settings}
To evaluate TPRNN, we conduct experiments on seven real-world time series forecasting datasets. Following previous work \cite{DLinear}, \textit{Electricity}, \textit{Traffic}, and \textit{Weather} datasets are divided into training, validation, and test splits in a ratio of 7:1:2 based on chronological order. \textit{ETTh1}, \textit{ETTh2}, \textit{ETTm1}, and \textit{ETTm2} datasets are divided into training, validation, and test splits in a ratio of 6:2:2 based on chronological order. Table \ref{tab:datasets} shows the statistical information of these datasets.
\begin{itemize}
    \item \textit{Electricity} dataset consists of hourly electricity consumption data, with recorded values representing the energy usage of 321 customers from 2012 to 2014.
    \item \textit{Traffic} dataset records the hourly traffic conditions of 862 road segments on the San Francisco Bay Highway from 2016 to 2018.
    \item \textit{Weather} dataset contains 21 environmental measurements recorded every 10 minutes throughout the year 2020.
    \item \textit{ETT} (Electricity Transformer Temperature) dataset collects two hourly-level datasets (i.e., \textit{ETTh1} and \textit{ETTh2}) and two 15-minute-level datasets (i.e., \textit{ETTm1} and \textit{ETTm2}), each of them contains 7 load features of electricity transformers from July 2016 to July 2018.
\end{itemize}
\begin{table}[]
\centering
\caption{The statistical information of seven real-world time series forecasting datasets.}
\label{tab:datasets}
\resizebox{0.55\columnwidth}{!}{
    \begin{tabular}{lccccc}
    \hline
    Datasets    & Electricity & Traffic & Weather & ETTh1\&2 & ETTm1\&2 \\ \hline
    Variates    & 321         & 862     & 21      & 7       & 7      \\
    Timesteps   & 26,304      & 17,544  & 52,696  & 17,420 & 69,680 \\
    Granularity & 1 hour      & 1 hour  & 10 min  & 1 hour & 15 min \\ \hline
    \end{tabular}
}
\end{table}

For experimental settings, TPRNN is implemented in Python 3.9 with PyTorch 1.11.0 and trained on a NVIDIA GeForce RTX 3090 GPU card. The optimizer of TPRNN is Adam, with an initial learning rate of 0.001, and the batch size is set to 32. To prevent overfitting during the model training process, we set the maximum number of epochs to 30. In addition, at the end of each epoch, we evaluate the performance using the validation set. If the current loss on the validation set exceeds the best loss for more than five consecutive times, we stop the training process and return the best model. The loss function is L1 loss function:
\begin{equation}
    \mathcal{L}(\theta) = \frac{1}{H}{\sum_{t = T + 1}^{T + H}\left\| {\bm{Y}_{t} - {\hat{\bm{Y}}}_{t}} \right\|}
\end{equation}
where $\theta$ denotes all the learnable parameters in TPRNN. $\bm{Y}_t$ and $\hat{\bm{Y}}$ are the ground truth and the forecasting results, respectively. We choose LSTM in the intra-scale interaction block and evaluate the performance of time series forecasting models using two metrics: Mean Squared Error (MSE) and Mean Absolute Error (MAE). The source code of TPRNN is released on GitHub\footnote{\url{https://github.com/cui-jia-hua/TPRNN}}.

\subsection{Methods for Comparison}

To validate the effectiveness of TPRNN, we compare it with six state-of-the-art time series forecasting models, including LogTrans \cite{LogTrans}, Informer \cite{Informer}, Autoformer \cite{Autoformer}, FEDformer \cite{FEDformer}, Pyraformer \cite{Pyraformer}, and DLinear \cite{DLinear}. To ensure a fair comparison, we adopt the same experimental settings and evaluation metrics used in \cite{FEDformer}. The detailed descriptions of baselines are as follows:

\begin{itemize}
  \item {LogTrans \cite{LogTrans}:} A LogSparse Transformer, which uses convolution and attention mechanisms to enhance the local information and reduce the computational complexity.
  \item {Informer \cite{Informer}:} An efficient Transformer, which uses a KL-divergence-based method to select dominant queries and reduce the computational complexity.
  \item {Autoformer \cite{Autoformer}:} A decomposition-based Transformer with auto-correlation, which uses a seasonal-trend decomposition architecture to split time series and an auto-correlation mechanisms instead of self-attention to focus on the connections of subsequences in time series.
  \item {FEDformer \cite{FEDformer}:} A frequency enhanced decomposed Transformer, which uses experts’ strategies to decompose time series and calculates attention value between different frequency components instead of different time points to model the global patterns in time series.
  \item {Pyraformer \cite{Pyraformer}:} A pyramid-structured Transformer, which extends the global-local structure of ETC [1], and calculates attention values between nodes and their neighbors (including neighboring nodes, parent node, and child nodes).
  \item {DLinear \cite{DLinear}:} A decomposition-based neural network, which uses linear layers to decompose the input into seasonal and trend components and employs fully connected layers to model each component separately.
\end{itemize}

\begin{table*}[]
    \centering
    \caption{The results of all models on seven datasets. The input length is set to 96 and the prediction length is set to 96, 192, 336, and 720. The best results are bolded and the second best results are underlined. IMP shows the improvement of TPRNN over the best baseline.}
    \label{tab:main_result}
    \resizebox{.95\textwidth}{!}{
    \begin{tabular}{cccccccccccccccccc}
\hline
\multicolumn{2}{c}{Models} & \multicolumn{2}{c}{\begin{tabular}[c]{@{}c@{}}TPRNN    \\ (ours)\end{tabular}} & \multicolumn{2}{c}{\begin{tabular}[c]{@{}c@{}}DLinear*    \\ (AAAI 2023)\end{tabular}} & \multicolumn{2}{c}{\begin{tabular}[c]{@{}c@{}}Pyraformer*\\ (ICLR 2022)\end{tabular}} & \multicolumn{2}{c}{\begin{tabular}[c]{@{}c@{}}FEDformer\\ (ICML 2022)\end{tabular}} & \multicolumn{2}{c}{\begin{tabular}[c]{@{}c@{}}Autoformer\\ (NIPS 2021)\end{tabular}} & \multicolumn{2}{c}{\begin{tabular}[c]{@{}c@{}}Informer\\ (AAAI 2021)\end{tabular}} & \multicolumn{2}{c}{\begin{tabular}[c]{@{}c@{}}LogTrans\\ (NIPS 2019)\end{tabular}} & \multicolumn{2}{c}{IMP} \\ \hline
\multicolumn{2}{c|}{Metrics} & MSE & MAE & MSE & MAE & MSE & MAE & MSE & MAE & MSE & MAE & MSE & MAE & MSE & MAE & MSE & MAE \\ \hline
\multicolumn{1}{c|}{\multirow{4}{*}{\rotatebox{90}{Electricity}}} & \multicolumn{1}{c|}{96} & \textbf{0.149} & \textbf{0.254} & 0.194 & \underline{0.276} & 0.386 & 0.449 & \underline{0.193} & 0.308 & 0.201 & 0.317 & 0.274 & 0.368 & 0.258 & 0.357 & 22.79\% & 7.97\% \\
\multicolumn{1}{c|}{} & \multicolumn{1}{c|}{192} & \textbf{0.167} & \textbf{0.274} & \underline{0.193} & \underline{0.280} & 0.378 & 0.443 & 0.201 & 0.315 & 0.222 & 0.334 & 0.296 & 0.386 & 0.266 & 0.368 & 13.47\% & 2.14\% \\
\multicolumn{1}{c|}{} & \multicolumn{1}{c|}{336} & \textbf{0.181} & \textbf{0.291} & \underline{0.205} & \underline{0.295} & 0.376 & 0.443 & 0.214 & 0.329 & 0.231 & 0.338 & 0.300 & 0.394 & 0.280 & 0.380 & 11.70\% & 1.35\% \\
\multicolumn{1}{c|}{} & \multicolumn{1}{c|}{720} & \textbf{0.207} & \textbf{0.313} & \underline{0.242} & \underline{0.329} & 0.376 & 0.445 & 0.246 & 0.355 & 0.254 & 0.361 & 0.373 & 0.439 & 0.283 & 0.376 & 14.46\% & 4.86\% \\ \hline
\multicolumn{1}{c|}{\multirow{4}{*}{\rotatebox{90}{Traffic}}} & \multicolumn{1}{c|}{96} & \textbf{0.560} & \textbf{0.354} & 0.652 & 0.398 & 0.867 & 0.468 & \underline{0.587} & \underline{0.366} & 0.613 & 0.388 & 0.719 & 0.391 & 0.684 & 0.384 & 4.60\% & 3.27\% \\
\multicolumn{1}{c|}{} & \multicolumn{1}{c|}{192} & \textbf{0.562} & \textbf{0.356} & \underline{0.598} & \underline{0.371} & 0.869 & 0.467 & 0.604 & 0.373 & 0.616 & 0.382 & 0.696 & 0.379 & 0.685 & 0.390 & 6.02\% & 4.04\% \\
\multicolumn{1}{c|}{} & \multicolumn{1}{c|}{336} & \textbf{0.577} & \underline{0.356} & \underline{0.606} & 0.374 & 0.881 & 0.469 & 0.621 & 0.383 & 0.622 & \textbf{0.337} & 0.777 & 0.420 & 0.734 & 0.408 & 4.78\% & -5.63\% \\
\multicolumn{1}{c|}{} & \multicolumn{1}{c|}{720} & \textbf{0.606} & \textbf{0.367} & 0.645 & 0.394 & 0.896 & 0.473 & \underline{0.626} & \underline{0.382} & 0.660 & 0.408 & 0.864 & 0.472 & 0.717 & 0.396 & 3.19\% & 3.92\% \\ \hline
\multicolumn{1}{c|}{\multirow{4}{*}{\rotatebox{90}{Weather}}} & \multicolumn{1}{c|}{96} & \textbf{0.165} & \textbf{0.236} & \underline{0.195} & \underline{0.254} & 0.622 & 0.556 & 0.217 & 0.296 & 0.266 & 0.336 & 0.300 & 0.384 & 0.458 & 0.490 & 15.38\% & 7.08\% \\
\multicolumn{1}{c|}{} & \multicolumn{1}{c|}{192} & \textbf{0.217} & \textbf{0.288} & \underline{0.235} & \underline{0.292} & 0.739 & 0.624 & 0.276 & 0.336 & 0.307 & 0.367 & 0.598 & 0.544 & 0.658 & 0.589 & 7.65\% & 1.36\% \\
\multicolumn{1}{c|}{} & \multicolumn{1}{c|}{336} & \textbf{0.278} & \underline{0.337} & \underline{0.282} & \textbf{0.333} & 1.004 & 0.753 & 0.339 & 0.380 & 0.359 & 0.395 & 0.578 & 0.523 & 0.797 & 0.652 & 1.41\% & -1.20\% \\
\multicolumn{1}{c|}{} & \multicolumn{1}{c|}{720} & \underline{0.351} & \underline{0.385} & \textbf{0.344} & \textbf{0.381} & 1.420 & 0.934 & 0.403 & 0.428 & 0.419 & 0.428 & 1.059 & 0.741 & 0.869 & 0.675 & -2.03\% & -1.05\% \\ \hline
\multicolumn{1}{c|}{\multirow{4}{*}{\rotatebox{90}{ETTh1}}} & \multicolumn{1}{c|}{96} & \textbf{0.372} & \underline{0.405} & 0.383 & \textbf{0.397} & 0.664 & 0.612 & \underline{0.376} & 0.419 & 0.449 & 0.459 & 0.865 & 0.713 & 0.878 & 0.740 & 1.06\% & -2.01\% \\
\multicolumn{1}{c|}{} & \multicolumn{1}{c|}{192} & \textbf{0.418} & \underline{0.439} & 0.434 & \textbf{0.429} & 0.790 & 0.681 & \underline{0.420} & 0.448 & 0.500 & 0.482 & 1.008 & 0.792 & 1.037 & 0.824 & 0.47\% & -2.33\% \\
\multicolumn{1}{c|}{} & \multicolumn{1}{c|}{336} & \textbf{0.442} & \textbf{0.455} & 0.481 & \underline{0.459} & 0.891 & 0.738 & \underline{0.459} & 0.465 & 0.521 & 0.496 & 1.107 & 0.809 & 1.238 & 0.932 & 3.70\% & 0.87\% \\
\multicolumn{1}{c|}{} & \multicolumn{1}{c|}{720} & \textbf{0.466} & \textbf{0.488} & \underline{0.506} & \underline{0.503} & 0.963 & 0.782 & 0.506 & 0.507 & 0.514 & 0.512 & 1.181 & 0.865 & 1.135 & 0.852 & 7.90\% & 2.98\% \\ \hline
\multicolumn{1}{c|}{\multirow{4}{*}{\rotatebox{90}{ETTh2}}} & \multicolumn{1}{c|}{96} & \textbf{0.294} & \textbf{0.358} & \underline{0.323} & \underline{0.379} & 0.645 & 0.597 & 0.346 & 0.388 & 0.358 & 0.397 & 3.755 & 1.525 & 2.116 & 1.197 & 8.97\% & 5.54\% \\
\multicolumn{1}{c|}{} & \multicolumn{1}{c|}{192} & \textbf{0.376} & \textbf{0.410} & 0.440 & 0.452 & 0.788 & 0.683 & \underline{0.429} & \underline{0.439} & 0.456 & 0.452 & 5.602 & 1.931 & 4.315 & 1.635 & 12.35\% & 6.60\% \\
\multicolumn{1}{c|}{} & \multicolumn{1}{c|}{336} & \textbf{0.414} & \textbf{0.446} & 0.559 & 0.521 & 0.907 & 0.747 & 0.496 & 0.487 & \underline{0.482} & \underline{0.486} & 4.721 & 1.835 & 1.124 & 1.604 & 14.10\% & 8.23\% \\
\multicolumn{1}{c|}{} & \multicolumn{1}{c|}{720} & \textbf{0.412} & \textbf{0.457} & 0.789 & 0.639 & 0.963 & 0.783 & \underline{0.463} & \underline{0.474} & 0.515 & 0.511 & 3.647 & 1.625 & 3.188 & 1.540 & 11.01\% & 3.58\% \\ \hline
\multicolumn{1}{c|}{\multirow{4}{*}{\rotatebox{90}{ETTm1}}} & \multicolumn{1}{c|}{96} & \textbf{0.310} & \textbf{0.359} & \underline{0.345} & \underline{0.370} & 0.543 & 0.510 & 0.379 & 0.419 & 0.505 & 0.475 & 0.672 & 0.571 & 0.600 & 0.546 & 10.14\% & 2.97\% \\
\multicolumn{1}{c|}{} & \multicolumn{1}{c|}{192} & \textbf{0.352} & \textbf{0.387} & \underline{0.380} & \underline{0.390} & 0.557 & 0.537 & 0.426 & 0.441 & 0.553 & 0.496 & 0.795 & 0.669 & 0.837 & 0.700 & 7.36\% & 0.76\% \\
\multicolumn{1}{c|}{} & \multicolumn{1}{c|}{336} & \textbf{0.381} & \textbf{0.409} & \underline{0.412} & \underline{0.413} & 0.754 & 0.655 & 0.445 & 0.459 & 0.621 & 0.537 & 1.212 & 0.871 & 1.124 & 0.832 & 7.52\% & 0.96\% \\
\multicolumn{1}{c|}{} & \multicolumn{1}{c|}{720} & \textbf{0.428} & \textbf{0.442} & \underline{0.472} & \underline{0.451} & 0.908 & 0.724 & 0.543 & 0.490 & 0.671 & 0.561 & 1.166 & 0.823 & 1.153 & 0.820 & 9.32\% & 1.99\% \\ \hline
\multicolumn{1}{c|}{\multirow{4}{*}{\rotatebox{90}{ETTm2}}} & \multicolumn{1}{c|}{96} & \textbf{0.182} & \textbf{0.275} & \underline{0.185} & \underline{0.281} & 0.435 & 0.507 & 0.203 & 0.287 & 0.255 & 0.339 & 0.365 & 0.453 & 0.768 & 0.642 & 1.62\% & 2.13\% \\
\multicolumn{1}{c|}{} & \multicolumn{1}{c|}{192} & \textbf{0.244} & \textbf{0.311} & \underline{0.268} & 0.345 & 0.730 & 0.673 & 0.269 & \underline{0.328} & 0.281 & 0.340 & 0.533 & 0.563 & 0.989 & 0.757 & 8.95\% & 5.18\% \\
\multicolumn{1}{c|}{} & \multicolumn{1}{c|}{336} & \textbf{0.298} & \textbf{0.348} & 0.389 & 0.432 & 1.201 & 0.845 & \underline{0.325} & \underline{0.366} & 0.339 & 0.372 & 1.363 & 0.887 & 1.334 & 0.872 & 8.30\% & 4.91\% \\
\multicolumn{1}{c|}{} & \multicolumn{1}{c|}{720} & \textbf{0.373} & \textbf{0.400} & 0.557 & 0.527 & 3.625 & 1.451 & \underline{0.421} & \underline{0.415} & 0.433 & 0.432 & 3.379 & 1.338 & 3.048 & 1.328 & 11.40\% & 3.61\% \\ \hline
\end{tabular}
}
\begin{tablenotes}
\item{We refer to the baseline results from \cite{DLinear}. * denotes re-trained models based on specified input and output lengths.}
\end{tablenotes}
\end{table*}

\subsection{Main Results} 
Table \ref{tab:main_result} shows the results of all models on seven datasets. The following tendencies can be discerned:
\begin{itemize}
    \item Our model (TPRNN) outperforms all baseline models in most cases. Specifically, TPRNN has achieved comprehensive improvements over baseline models in time series forecasting. Compared to the best baseline model, TPRNN achieves average MSE improvements of 15.60\%, 4.65\%, 5.60\%, and 7.76\% on \textit{Electricity}, \textit{Traffic}, \textit{Weather}, and \textit{ETT} datasets, respectively. For datasets like \textit{Electricity} and \textit{ETT}, which show strong periodicity, TPRNN can effectively capture the underlying multi-scale temporal patterns, as evidenced by higher improvements compared to the best baseline.
    \item Among the baselines, the best-performing transformer-based model is FEDformer. Although transformer-based models can capture long-term dependencies, their design with self-attention mechanism does not allow for capturing the specific positional relationships of values. FEDformer addresses this issue by transforming the time series into the frequency domain and computing attention values of different frequency components, which considers the relative positions of values within the sequence, leading to superior forecasting performance compared to other transformer-based models.
    \item Pyraformer, which also models the multi-scale temporal patterns of time series, exhibits lower forecasting performance compared to TPRNN. While Pyraformer utilizes a pyramid attention module to capture the multi-scale temporal patterns, its capturing range is limited to only the neighboring nodes of the target node, which fails to capture dependencies between more distant nodes and the target node. TPRNN utilizes the intra-scale interaction block and the inter-scale interaction block in an alternating fashion, sequentially modeling subsequences of different scales from top to bottom and comprehensively capturing the influences of nodes of different scales.
\end{itemize}

\subsection{Effect of Mixed Multi-Scale Construction Module}
To demonstrate the effect of mixed multi-scale construction module, we evaluate the performance of TPRNN with the following two variants.
\begin{itemize}
    \item {TPRNN w/o conv:} This variant model removes convolution method from CSICB, and extracts feature from the data through three pooling methods (max pooling, min pooling, and average pooling).
    \item {TPRNN w/o pooling:} This variant model removes pooling methods from CSICB, and extracts feature from the data solely through convolution method.
\end{itemize}

\begin{table}[]
\caption{The results of different multi-scale construction methods. The best results are bolded.}
    \label{tab:table3}
    \resizebox{.45\textwidth}{!}{
\begin{tabular}{lccccc}
\hline
\multirow{2}{*}{Methods} & \multirow{2}{*}{Metrics}                          & \multicolumn{2}{c}{Electricity}                                                                                                   & \multicolumn{2}{c}{ETTh1}                                                                                                         \\ \cline{3-6} 
                         &                                                   & \multicolumn{1}{c}{96}                                                    & 720                                                   & \multicolumn{1}{c}{96}                                                    & 720                                                   \\ \hline
TPRNN w/o conv           & \begin{tabular}[c]{@{}c@{}}MSE\\ MAE\end{tabular} & \multicolumn{1}{c}{\begin{tabular}[c]{@{}c@{}}0.151\\ 0.258\end{tabular}} & \begin{tabular}[c]{@{}c@{}}0.225\\ 0.327\end{tabular} & \multicolumn{1}{c}{\begin{tabular}[c]{@{}c@{}}0.383\\ 0.407\end{tabular}} & \begin{tabular}[c]{@{}c@{}}0.509\\ 0.515\end{tabular} \\ \hline
TPRNN w/o pooling        & \begin{tabular}[c]{@{}c@{}}MSE\\ MAE\end{tabular} & \multicolumn{1}{c}{\begin{tabular}[c]{@{}c@{}}0.152\\ 0.259\end{tabular}} & \begin{tabular}[c]{@{}c@{}}0.215\\ 0.319\end{tabular} & \multicolumn{1}{c}{\begin{tabular}[c]{@{}c@{}}0.390\\ 0.414\end{tabular}} & \begin{tabular}[c]{@{}c@{}}0.498\\ 0.510\end{tabular} \\ \hline
TPRNN                    & \begin{tabular}[c]{@{}c@{}}MSE\\ MAE\end{tabular} & \multicolumn{1}{c}{\begin{tabular}[c]{@{}c@{}}\textbf{0.149}\\ \textbf{0.254}\end{tabular}} & \begin{tabular}[c]{@{}c@{}}\textbf{0.210}\\ \textbf{0.318}\end{tabular} & \multicolumn{1}{c}{\begin{tabular}[c]{@{}c@{}}\textbf{0.372}\\ \textbf{0.405}\end{tabular}} & \begin{tabular}[c]{@{}c@{}}\textbf{0.479}\\ \textbf{0.500}\end{tabular} \\ \hline
\end{tabular}
}
\end{table}
From the results in Table \ref{tab:table3}, we can see that using both convolution and pooling methods to obtain subsequences of different scales leads to more accurate results. Furthermore, it can be observed that when the prediction length is short (i.e., 96), using only the convolution-based subsequences are not as effective as using only the pooling-based subsequences. However, when the prediction length is long (i.e., 720), the opposite is true. This indicates that the pooling-based construction method retains more details of the input, making it more suitable for short-term predictions and the convolution-based construction method excels in extracting global features, making it more suitable for long-term predictions.

\subsection{Effect of Multi-Scale Information Interaction Module} To demonstrate the effect of multi-scale information interaction module, we evaluate the performance of TPRNN with the following five variants.

\begin{itemize}
    \item {TPRNN w/o all:} This variant model removes the multi-scale information interaction module from TPRNN. After the mixed multi-scale construction module, subsequences of different scales are directly input into the fusion prediction module for forecasting.
    \item {TPRNN w/o interscale:} This variant model removes the inter-scale interaction block from the multi-scale information interaction module. There is no interaction of subsequences of different scales, and the subsequences obtained from the intra-scale interaction block are directly input into the fusion prediction module for forecasting.
    \item {TPRNN w/o intrascale:} This variant model removes the intra-scale interaction block from the multi-scale information interaction module, which directly uses the inter-scale interaction block to model the interactions of subsequences, without relying on RNN to model the temporal dependencies within each scale.
    \item {TPRNN lastnode:} This variant model treats the last node of the subsequence modeled by RNN as the global information and directly fuses the global information with the adjacent smaller-scale subsequences using a linear layer, eliminating the computation of global information.
    \item {TPRNN fullconnect:} This variant model uses one linear layer to directly map the subsequence to the length of the adjacent smaller-scale subsequence, computing the interactions of all nodes between two subsequences.
\end{itemize}

\begin{table}[]
\caption{The results of different information interaction methods. The best results are bolded.}
    \label{tab:table4}
    \resizebox{.45\textwidth}{!}{
\begin{tabular}{lccccc}
\hline
\multirow{2}{*}{Models} & \multirow{2}{*}{Metrics}                          & \multicolumn{2}{c}{Electricity}                                                                                                                     & \multicolumn{2}{c}{ETTh1}                                                                                                                           \\ \cline{3-6} 
                        &                                                   & \multicolumn{1}{c}{96}                                                             & 720                                                            & \multicolumn{1}{c}{96}                                                             & 720                                                            \\ \hline
TPRNN w/o all           & \begin{tabular}[c]{@{}c@{}}MSE\\ MAE\end{tabular} & \multicolumn{1}{c}{\begin{tabular}[c]{@{}c@{}}0.195\\ 0.303\end{tabular}}          & \begin{tabular}[c]{@{}c@{}}0.253\\ 0.348\end{tabular}          & \multicolumn{1}{c}{\begin{tabular}[c]{@{}c@{}}0.382\\ 0.396\end{tabular}}          & \begin{tabular}[c]{@{}c@{}}0.489\\ 0.472\end{tabular}          \\ \hline
TPRNN w/o interscale    & \begin{tabular}[c]{@{}c@{}}MSE\\ MAE\end{tabular} & \multicolumn{1}{c}{\begin{tabular}[c]{@{}c@{}}0.154\\ 0.259\end{tabular}}          & \begin{tabular}[c]{@{}c@{}}0.220\\ 0.323\end{tabular}          & \multicolumn{1}{c}{\begin{tabular}[c]{@{}c@{}}0.384\\ 0.409\end{tabular}}          & \begin{tabular}[c]{@{}c@{}}0.487\\ 0.507\end{tabular}          \\ \hline
TPRNN w/o intrascale    & \begin{tabular}[c]{@{}c@{}}MSE\\ MAE\end{tabular} & \multicolumn{1}{c}{\begin{tabular}[c]{@{}c@{}}0.169\\ 0.283\end{tabular}}          & \begin{tabular}[c]{@{}c@{}}0.220\\ 0.327\end{tabular}          & \multicolumn{1}{c}{\begin{tabular}[c]{@{}c@{}}0.386\\ 0.410\end{tabular}}          & \begin{tabular}[c]{@{}c@{}}0.508\\ 0.498\end{tabular}          \\ \hline
TPRNN lastnode          & \begin{tabular}[c]{@{}c@{}}MSE\\ MAE\end{tabular} & \multicolumn{1}{c}{\begin{tabular}[c]{@{}c@{}}0.173\\ 0.286\end{tabular}}          & \begin{tabular}[c]{@{}c@{}}0.221\\ 0.328\end{tabular}          & \multicolumn{1}{c}{\begin{tabular}[c]{@{}c@{}}0.394\\ 0.412\end{tabular}}          & \begin{tabular}[c]{@{}c@{}}0.519\\ 0.519\end{tabular}          \\ \hline
TPRNN fullconnect       & \begin{tabular}[c]{@{}c@{}}MSE\\ MAE\end{tabular} & \multicolumn{1}{c}{\begin{tabular}[c]{@{}c@{}}0.157\\ 0.264\end{tabular}}          & \begin{tabular}[c]{@{}c@{}}0.212\\ 0.319\end{tabular}          & \multicolumn{1}{c}{\begin{tabular}[c]{@{}c@{}}0.383\\ 0.414\end{tabular}}          & \begin{tabular}[c]{@{}c@{}}0.486\\ 0.502\end{tabular}          \\ \hline
TPRNN                   & \begin{tabular}[c]{@{}c@{}}MSE\\ MAE\end{tabular} & \multicolumn{1}{c}{\textbf{\begin{tabular}[c]{@{}c@{}}0.149\\ 0.254\end{tabular}}} & \textbf{\begin{tabular}[c]{@{}c@{}}0.210\\ 0.318\end{tabular}} & \multicolumn{1}{c}{\textbf{\begin{tabular}[c]{@{}c@{}}0.372\\ 0.405\end{tabular}}} & \textbf{\begin{tabular}[c]{@{}c@{}}0.479\\ 0.500\end{tabular}} \\ \hline
\end{tabular}
}
\end{table}

From the results in Table \ref{tab:table4}, we can see that the inter-scale interaction block and the intra-scale interaction block complement each other. Removing either block would significantly decrease the accuracy of predictions. In addition, it is evident that utilizing the last node of the RNN as global information, as done in “TPRNN lastnode”, fails to achieve accurate predictions. This is due to the vanishing gradient problem of RNNs, where the last node only receives recent information while ignoring distant historical information. In most scenarios, directly calculating the influences between subsequences through a linear layer slightly decreases the forecasting performance compared to the inter-scale interaction block in TPRNN. This is because directly calculating the impact of nodes of different layers results in a higher number of parameters, leading to overfitting.

\subsection{Effect of Fusion Prediction Module} To demonstrate the effect of multi-scale information interaction module, we evaluate the performance of TPRNN with the following variant.
\begin{itemize}
    \item {TPRNN w/o fusion:} This variant model removes the fusion prediction module from TPRNN. After the multi-scale information interaction module, only the subsequence of the original scale is used for prediction, without employing predictors from other scales.
\end{itemize}

\begin{table}[]
    \caption{The results of different fusion methods. The best results are bolded.}
    \label{tab:table5}
    \resizebox{.45\textwidth}{!}{
\begin{tabular}{lccccc}
\hline
\multirow{2}{*}{Models} & \multirow{2}{*}{Metrics}                          & \multicolumn{2}{c}{Electricity}                                                                                                                     & \multicolumn{2}{c}{ETTh1}                                                                                                                           \\ \cline{3-6} 
                        &                                                   & \multicolumn{1}{c}{96}                                                             & 720                                                            & \multicolumn{1}{c}{96}                                                             & 720                                                            \\ \hline
TPRNN w/o fusion        & \begin{tabular}[c]{@{}c@{}}MSE\\ MAE\end{tabular} & \multicolumn{1}{c}{\begin{tabular}[c]{@{}c@{}}0.193\\ 0.308\end{tabular}}          & \begin{tabular}[c]{@{}c@{}}0.218\\ 0.326\end{tabular}          & \multicolumn{1}{c}{\begin{tabular}[c]{@{}c@{}}0.389\\ 0.407\end{tabular}}          & \textbf{\begin{tabular}[c]{@{}c@{}}0.478\\ 0.486\end{tabular}} \\ \hline
TPRNN                   & \begin{tabular}[c]{@{}c@{}}MSE\\ MAE\end{tabular} & \multicolumn{1}{c}{\textbf{\begin{tabular}[c]{@{}c@{}}0.149\\ 0.254\end{tabular}}} & \textbf{\begin{tabular}[c]{@{}c@{}}0.210\\ 0.318\end{tabular}} & \multicolumn{1}{c}{\textbf{\begin{tabular}[c]{@{}c@{}}0.372\\ 0.405\end{tabular}}} & \begin{tabular}[c]{@{}c@{}}0.479\\ 0.500\end{tabular}          \\ \hline
\end{tabular}
}
\end{table}

From the results in Table \ref{tab:table5}, we can see that removing the fusion prediction module generally leads to a decrease in forecasting performance. However, in the case of long-term forecasting on \textit{ETTh1} dataset, the non-fusion prediction performs slightly better than the fusion prediction. The reason is that \textit{ETTh1} dataset exhibits strong periodicity, and accurate predictions can be achieved without considering large-scale information. Due to the introduction of additional parameters, the fusion prediction module might lead to overfitting. On \textit{Electricity} dataset, which exhibits more complex temporal patterns, the utilization of the fusion prediction module obtains significant performance improvement.

\begin{figure}[t]
\centering
\includegraphics[width=0.85\columnwidth]{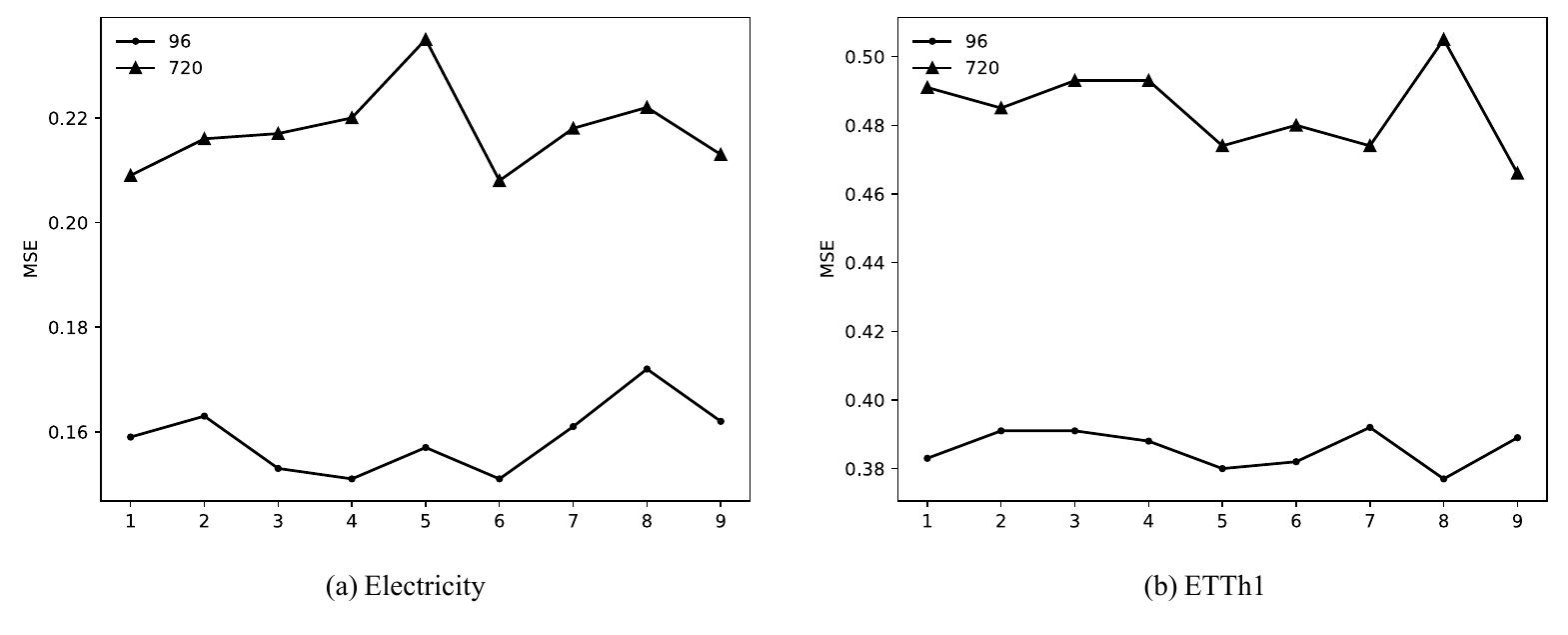}
\caption{The MSE result of TPRNN with varying global information length on \textit{Electricity} and \textit{ETTh1} datasets, where the input length and prediction length are both set to 96.}
\label{fig:hyperparameter}
\end{figure}

\subsection{Hyperparameter Analysis}
The length of global information in the inter-scale interaction block is an important hyperparameter in TPRNN. Fig. \ref{fig:hyperparameter} shows the MSE of TPRNN on \textit{Electricity} and \textit{ETTh1} datasets by varying global information length from 1 to 10. From the results, it can be observed that different lengths of global information significantly affect the model’s performance. Setting the global information length to 6 achieves the best average performance for TPRNN. This is because setting the global information length too large would rapidly increase the number of parameters, leading to overfitting and introducing noise. On the other hand, if the length is too small, the subsequence may not capture the full extent of global information, leading to inadequate information interaction.

\subsection{Model Analysis}
\begin{figure*}[t]
\centering
\includegraphics[width=0.85\textwidth]{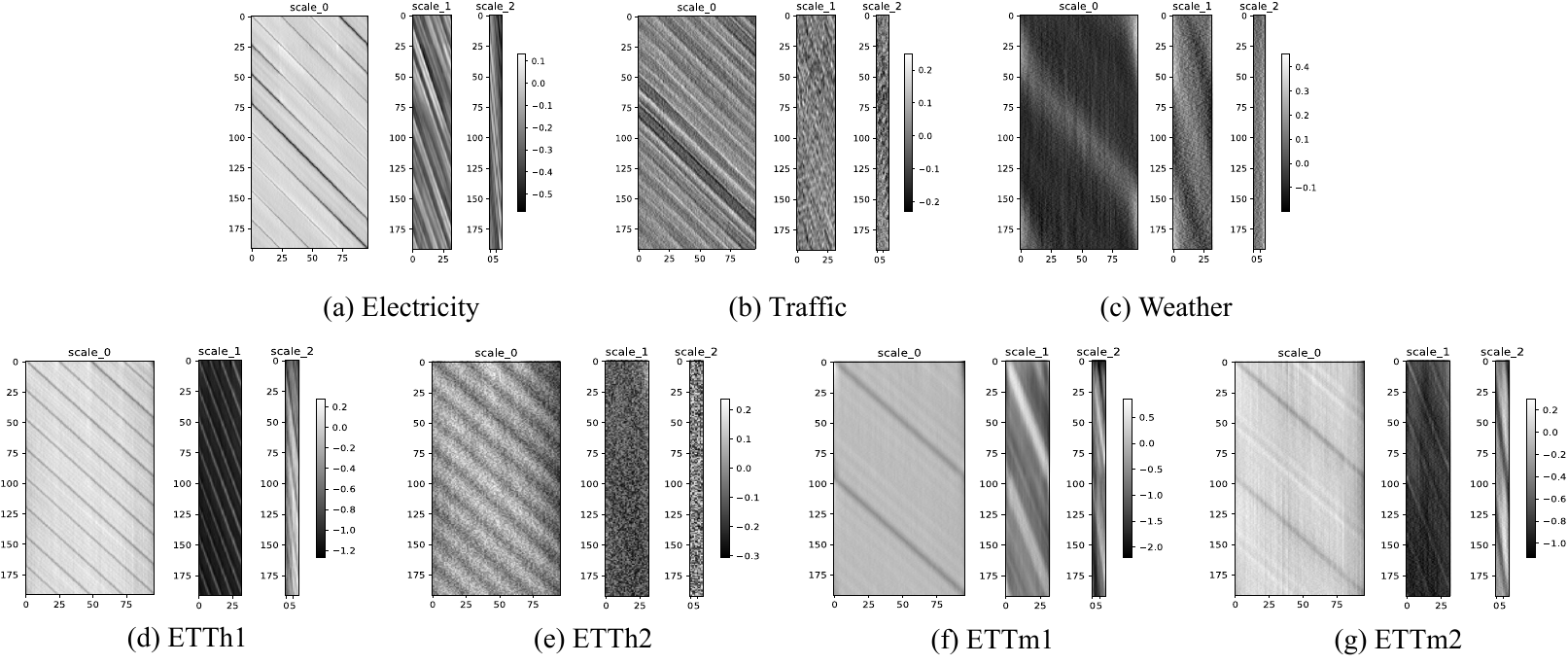}
\caption{The visualization of the predictor parameters in the fusion prediction module on seven datasets. We train TPRNN with a specific configuration of $C=2$, where $C$ represents the number of subsequences of different scales, with the input length set to 96 and the prediction length set to 192.}
\label{fig:model_ana}
\end{figure*}

The predictor in TPRNN is implemented by linear layers, so the parameters can effectively reflect the impact of different positions on the predictions. By visualizing the weights of the linear layer, as shown in Fig. \ref{fig:model_ana}, we can intuitively observe the influences of different positions within subsequences of different scales on the predictions. It can be observed that on \textit{Electricity}, \textit{Traffic}, and \textit{ETT} datasets (related to human activities), there are noticeable patterns in the contribution of positions within subsequences to the prediction. For \textit{Electricity}, \textit{Traffic}, \textit{ETTh1}, and \textit{ETTh2} datasets, which have a sampling rate of one hour, the cycle of these patterns is 24. For \textit{ETTm1} and \textit{ETTm2} datasets, which have a sampling rate of 15 minutes, the cycle is 96. However, for \textit{Weather} dataset, where the data changes are not directly related to human activities, it is challenging to observe a clear cycle. In addition, it can be observed that the patterns in different subsequences are similar to each other, with differences mainly in the scale.

\section{Conclusions and future work}
In this paper, we propose a Top-down Pyramidal Recurrent Neural Network (TPRNN) for time series forecasting. Specifically, the mixed multi-scale construction module is introduced to form a pyramid structure of subsequences of different scales. The multi-scale information interaction module, which consists of the intra-scale interaction block and the inter-scale interaction block, is introduced to model both the temporal dependencies in each scale and the influences of subsequences of different scales in a top-down manner within the pyramid structure. In addition, the fusion prediction module is introduced to combine the predictions of different scales. TPRNN achieves competitive performance on seven real-world datasets, outperforming the best baseline with an average improvement of 8.13\% in MSE.

In the future, this work can be extended in the following directions. First, we will introduce a weight update strategy to adaptively adjust weights at different scales for different datasets, which can enhance the utilization of multi-scale subsequences. Second, we will design a mixture multi-scale graph to capture interactions between subsequences at all scales, which can enhance the modeling ability for temporal dependencies.

\bibliographystyle{ACM-Reference-Format}
\bibliography{TPRNN_fin}

\end{document}